\documentclass[runningheads]{llncs}
\usepackage{graphicx}
\usepackage{amsmath,amssymb} 
\usepackage{color}
\usepackage[width=122mm,left=12mm,paperwidth=146mm,height=193mm,top=12mm,paperheight=217mm]{geometry}
\usepackage[colorlinks=true,citecolor=green]{hyperref}
\usepackage{hhline}
\usepackage{multicol}
\usepackage{changepage}
\usepackage{overpic}
\usepackage{multicol}
\usepackage{url}

\newcommand\ignore[1]{}

\newcommand{\eg}{\textit{e.g.}}
\newcommand{\ie}{\textit{i.e.}}
\newcommand{\etal}{\textit{et al.}}

\DeclareMathOperator*{\argmin}{argmin}
\newcommand{\norm}[1]{\left\lVert#1\right\rVert}

\def \path{\bp C}

\newcommand{\bfd}{{\mathbf{d}}}

\newcommand{\bfx}{{\mathbf{x}}}

\newcommand{\bfM}{{\mathbf{M}}}

\begin{document}
\pagestyle{headings}
\mainmatter
\def\ECCV16SubNumber{}  

\title{Learning how to be robust: \\Deep polynomial regression} 

\titlerunning{deep polynomial regression \ECCV16SubNumber}

\authorrunning{P\'erez-R\'ua \etal}

\author{Juan-Manuel P\'erez-R\'ua$^{1,2}$\thanks{\scriptsize Now with Orange Labs, France}\and Tomas Crivelli$^{1}$\thanks{\scriptsize Now with Zowl Labs, Argentina} \and \newline Patrick Bouthemy$^{2}$\and Patrick P\'erez$^{1}$\thanks{\scriptsize Now with Valeo.ai, France}}

\institute{$^1$ Technicolor, Cesson S\'evign\'e, France \newline $^2$ Inria, Centre Rennes -- Bretagne Atlantique, France}

\maketitle

\begin{abstract}

Polynomial regression is a recurrent problem with a large number of applications. In computer vision it often appears in motion analysis. Whatever the application, standard methods for regression of polynomial models tend to deliver biased results when the input data is heavily contaminated by outliers. Moreover, the problem is even harder when outliers have strong structure. Departing from problem-tailored heuristics for robust estimation of parametric models, we explore deep convolutional neural networks. Our work aims to find a generic approach for training deep regression models without the explicit need of supervised annotation. We bypass the need for a tailored loss function on the regression parameters by attaching to our model a differentiable hard-wired decoder corresponding to the polynomial operation at hand. We demonstrate the value of our findings by comparing with standard robust regression methods. Furthermore, we demonstrate how to use such models for a real computer vision problem, \ie, video stabilization. The qualitative and quantitative experiments show that neural networks are able to learn robustness for general polynomial regression, with results that well overpass scores of traditional robust estimation methods.

\keywords{Deep learning, polynomial regression, parameric motion model}
\end{abstract}

\section{Introduction}
\label{sec:intro}

Fitting a finite degree polynomial model to a set of measurements is a problem that appears recurrently in machine learning and computer vision~\cite{meer1991robust}. It is known as polynomial fitting or polynomial regression. When the input data follow one instance of the model class, exactly or up to an additive white Gaussian noise,

the optimal estimator of the polynomial coefficients\footnote{Conventionally, in the deep learning literature we call ``parameters'' the set of values that are learned during training (connection weights essentially). Sometimes, the word ``parameters'' also refers to the coefficients of a regressed polynomial. To avoid confusion for  the latter meaning, we use either the word ``coefficients'' in the first part of this manuscript or the phrase ``parametric motion model'' in the second part.} is the least squares estimator (LSE). However, in very few domains one would encounter such a situation. In reality, data is usually affected not only by noise, but by non-trivial interference, blind spots (unmasked missing data), and  many other types of outliers. In these scenarios, LSE is biased.  

Attempts to account for the wide variety of input data contamination, including structured outliers, have been proposed in the past. These include specific heuristics like random sample consensus~\cite{fischler1981random} (RANSAC) or one of its many problem-specific variations. Robust statistics have enjoyed popularity among researchers as well. However, these solutions sometimes require a great deal of tuning, while still leaving room for improvement on the estimation accuracy and insensitivity to structured outliers. Moreover, most of the available techniques for robust estimation rely on specific priors on the input data, for instance, expected ratio of outliers~\cite{fischler1981random} or rough localization of them, as it is expressed by alternate optimization in~\cite{puy2014robust}. It is precisely with the goal of eliminating as much as possible any need for prior knowledge on the input data that we explore deep neural networks in this context. We hypothesize that the multi-scale spatial reasoning of a model empowered with stacks of convolutional layers is key towards universally robust polynomial regression. 

Indeed, deep models were found to be useful in a large variety of complex regression problems~\cite{belagiannis2015robust,tewari2017mofa,newell2016stacked}. The ubiquity of convolutional neural networks in these type of problems speaks of their potential for the task of polynomial regression. A particular property we are specially interested in this paper is robustness and how to learn to be robust. However, during supervised learning, the types of robustness a model can learn are tightly related to the examples from the training dataset. Given the difficulties that arise during collecting the large datasets that neural networks need, it is very likely that for a given problem only a small portion of those cases are covered. How to help deep models generalize for other cases is an open question. In practice, this is usually handled by randomized data augmentation~\cite{ciregan2012multi}. Indeed, being able to generalize from the training dataset, and being robust to damaged input seem to be, at least in principle, related concepts in machine learning.

Another difficult question that arises when training such models for regression problems is what is the best loss function. In particular when regressing coefficients of a polynomial function, standard loss functions might not be optimal. This is related with the fact that, very often for some problems, few coefficients are much larger than others, causing imbalance during training. This might be the reason why for optical flow, a common regression problem in computer vision~\cite{dosovitskiy2015flownet}, a convolutional neural network trained with the $L2$-loss learns much more easily to predict large motion vectors than smaller ones.

With all these ideas in mind, the main contributions of this work can be summarized as follows:

\begin{itemize}
	\item Describing a family of deep models for polynomial regression,
	\item Defining a simple methodology for unsupervised training of polynomial regression models,
	\item Comparing the effect of a loss function applied on the output data stemming from the estimated parametric model vs. a loss applied directly on the estimated polynomial coefficients,
	\item Exploring the effect of robust losses during training,
	\item Analyzing polynomial regression problems of different input data dimensionality.
	\item A simple application for estimation of parametric motion models and video stabilization.
\end{itemize}

We start by summarizing the related work in Section~\ref{sec:related}. Motivating ideas for our work are discussed in Section~\ref{sec:lessons}. We then explain our models in Section~\ref{sec:core}, and give way to the core experimental work in Section~\ref{sec:experiments}. Final comments are given in Section~\ref{sec:conclusions}.

\section{Related work}
\label{sec:related}

In this section we give a review of the related work. First, we start with a brief introduction to robust regression methods. We include a description and motivation of iterative methods like RANSAC and consensus-based approaches, and continue with robust estimators. Later, we explain further the problem of parametric motion model estimation, which is a form of regression often found in the computer vision literature. Finally, we introduce recent works on deep models for regression and similar tasks.

\subsection{Robust regression}

\textbf{RANSAC}, proposed by Fischler and Bolles in 1981~\cite{fischler1981random}, is an iterative method for alternated determination of model inliers and model parameters. It encompasses randomized sampling of the input data set, estimation of a candidate parametric model explaining the chosen subset, and determining the proportion of data points that agree with the candidate model by using a hand-tuned threshold. The method iterates for a fixed number of iterations or until enough data points find a consensus. The randomized nature of the method implies that for a single dataset results of multiple runs might be different. Furthermore, the algorithm parameters usually need to be tuned up for different problems, and it is known to be sensitive to the choice of the threshold~\cite{torr1997development}. To provide some more stability to this random heuristic, some works have focused on other ways to establish goodness of fit: least median squares, which even though it offers outstanding robustness, still fails when the ratio of outliers is very large; MLESAC, which maximizes the likelihood rather than number of inliers~\cite{torr2000mlesac}; MINPRAN~\cite{stewart1995minpran}, which makes assumptions on the randomness of the data, etc.

\textbf{Robust estimators} aim to fix the bias problem of LSE, by replacing the $L2$-norm with more conservative penalties when the residual error is large. A common idea is to replace the $L2$-norm with $L1$, but this change only increases robustness for the mono-dimensional case, where minimizing the $L1$-norm acts as estimator of the median. Truncated least-squares and least-trimmed squares are other options to replace $L2$-norm. Departing from these, more complex functions with certain desirable properties were proposed. Of considerable popularity in computer vision, redescending M-estimators offer certain theoretical advantages over other robust estimators. For instance, they mostly ignore large outliers~\cite{huber2011robust}. Generally speaking, however, there is no good algorithm for selecting one estimator from the variety of M-estimators. Very often is also difficult to solve for the selected robust penalty. A solution that works well in many cases is the Iteratively Re-weighted Least Squares algorithm~\cite{holland1977robust}.

\subsection{Parametric motion models}

We make a short overview of a common application of polynomial regression in computer vision: estimation of parametric motion models. This use-case is a great example of polynomial regression with strong outliers. Natural scenes can often be roughly separated into background and foreground segments. Foreground segments often include moving people, vehicles or any type of independently moving objects. When the task is estimating the dominant image motion due to camera movements, foreground segments can be effectively seen as spatially-coherent outliers. Depending on the scene, these outliers can occupy a very large portion of the image support, hindering accurate estimation.  

In many dynamic scene analysis building blocks, accuracy of polynomial regression is important, \eg, in motion segmentation~\cite{odobez1997separation,cremers2005motion}, optical flow estimation~\cite{black1996estimating,farneback2003two,fortun2015aggregation,yang2015dense}, detection of motion anomalies~\cite{perez2017detection}, and tracking~\cite{black1995tracking}. 
Classical methods pose parametric motion model estimation as an inverse problem that is solved through minimization of an energy functional~\cite{black1996robust,bergen1992hierarchical}. These methods leverage the motion constraint to form a data driven term encouraging motion parameters that minimize the displaced frame difference (DFD) between the input images. In contrast to per-pixel optical flow estimation, the estimation of parametric motion models is not an underdetermined problem. Indeed, the proposed models explain the image-based motion cues for all the image pixels at once (or a subset of them). Usually the number of observations, \ie\ pixel positions, is much greater than the number of parameters of the motion model, leading to stable solutions when no motion outliers are present in the scene.

However, under the presence of outliers, models that simply penalize the displaced frame intensity difference with the $L2$ norm encounter estimation accuracy problems. In order to overcome these issues, several methods~\cite{yang2015dense,odobez1995robust,black1996robust} propose to use different robust penalties. In the presence of large displacements, and strong camera motions, Odobez and Bouthemy~\cite{odobez1995robust} proposed a multi-resolution, incremental and robust scheme where simpler parametric motion models are estimated at coarser scales, and incrementally updated at finer ones. More recently, in order to cope with the aperture problem, a strategy to adapt the support of regions where motion is estimated is presented by~\cite{senst2012robust}.

\subsection{Deep learning for regression problems}

Convolutional Neural Networks (CNNs) have started to dominate Computer Vision problems that had been traditionally very complicated to address with learning-based methods. This is most probably  due to the higher-level features that the hierarchical CNN architectures are able to learn. One example of these problems is the estimation of optical flow. In scenes with large motion ambiguity, only semantic cues are able to recover the correct apparent motion~\cite{dosovitskiy2015flownet,thewlis2016fully,ilg2017flownet,bailer2017cnn,sun2017pwc}. This seems to be the reason why deep optical flow methods are currently dominating benchmarks.

In Dosovitskiy \etal\ \cite{dosovitskiy2015flownet}\footnote{The use of convolutions for optical flow has a longer history. For instance, Farneb{\"a}ck implemented his motion estimation method by means of separable convolutions in~\cite{farneback2000fast}. Weinzaepfel \etal,~\cite{weinzaepfel2013deepflow} rely on a large stack of patch-based convolutional responses. To the best of our knowledge, however, the method of Dosovitskiy \etal\ \cite{dosovitskiy2015flownet} is actually the first one to use learned convolutional filters to perform the mapping between images and motion fields.} convolutional filters successfully learn how to estimate two dimensional motion fields from pairs of successive video frames. A few elements of this approach, coined FlowNet, have to be considered when tackling similar tasks. Performing a complex transform from 2D maps (images) to same resolution 2D maps (optical flows) requires to capture high level features from data. In order that features can pick up global information at the total spatial extension of the input maps, they are implemented in a contractive fashion. Indeed, this is a very common practice in applied deep learning. A second part of the network must then take those features
and expand them so that they are able to restore the spatial resolution of the output. An encoder-decoder architecture comes easily to mind. However, special attention must be taken for the motion estimation problem. Indeed, optical flow networks must have good localization properties. \textbf{Forward skip connections} from contractive layers are connected through convolutions to the expanding part of FlowNet, alleviating the bad localization issue of deep networks and simple encoder-decoder architectures.

A problem closely related to polynomial regression, geometric matching, consists of finding a parametric transformation of the image grid, allowing the registration of input frames. Recently, Rocco \etal, \cite{rocco2017convolutional} proposed a neural network model that is capable of registering pairs of images that do not necessarily belong to the same image sequence. The target parametric transformations were affine and thin plate splines~\cite{bookstein1989principal}. In their model, the problem is divided into three tasks: symmetric feature extraction with a Siamese network initialized with VGG features~\cite{simonyan2014very}, a dense correlation layer similar  to the one used by FlowNet, and a regression layer, which infers the image grid transformation.

Another regression problem that has been recently tackled by CNNs is human pose estimation. Excellent results were obtained with the so-called \textbf{stacked hourglass networks} (SHN)~\cite{newell2016stacked}.

\section{Lessons from the state-of-the-art}
\label{sec:lessons}

The success of deep models on the complicated tasks described in Section~\ref{sec:related} motivates the exploration of deep models for learning how to robustly estimate parametric models. 

One interesting element of FlowNet~\cite{dosovitskiy2015flownet,ilg2017flownet} is that it was trained on a \textbf{synthetic dataset} called \textit{FlyingChairs}. The dataset contains around 25,000 images of chairs on background images extracted randomly from \textit{Flickr}. The backgrounds were assigned with a random rigid motion, and the foreground, composed of computer-generated chairs, with another one. A simple strategy for data augmentation allows the network to generalize from that dataset to real images.
The final results of FlowNet are impressive considering that the pipeline is learned in an end-to-end fashion with synthetic data, and, powered by modern GPUs, they are computed in almost real-time. The evolution of FlowNet, FlowNet 2.0~\cite{ilg2017flownet}, ranks very highly in optical flow benchmarks. Perhaps the element introduced by FlowNet 2.0 that is most relevant to this work is \textbf{curriculum learning.} One of the issues of the original FlowNet is the poor behaviour for small displacements. To tackle this, FlowNet 2.0 leverages a second synthetic dataset depicting more complex motions (and of smaller magnitude in average), coined \textit{Flying3DThings}. The optimal schedule for training was to first use \textit{FlyingChairs}, and then the \textit{Flying3DThings}. Apparently, a neural network is predisposed to learn more complex data priors, when already trained for simpler ones. We will test this hypothesis for our scenario later on.
	
Newell \etal~\cite{newell2016stacked} stated the human pose estimation problem as a dense map-to-map inference problem. The important elements that allow such networks to perform so well can be summarized as follows:

\begin{itemize}
	\item \textbf{Skip layers} with symmetric connection from the convolutional operations in the contractive part of the network, to the upsampling layers in the expansive part of the network. This particular design essentially allows the network to be aware of global and local information at every stage of the decoding part. A single module with this design properties is called an hourglass module.
	
	\item \textbf{Stacks.} Stacking hourglass modules seems to allow the SHN to perform repeated top-down, bottom-up operations that might be essential on capturing different aspects of the pose estimation problem at every module.
	
	\item \textbf{Residual connections.} The residual connections, as introduced by~\cite{he2016deep} allow very deep models to be properly trained. Each residual module is by-passed by an identity transformation that allows gradients flow freely through the network. A deeper understanding of residual learning can be obtained by looking at~\cite{he2016identity}.
	
	\item \textbf{Intermediate supervision.} SHN allows intermediate outputs to be used in the training loss. This procedure guarantees that each hourglass module learns something about the pose estimation problem, and further stabilizes the overall training.
	
\end{itemize}

The lessons obtained from the start-of-the-art are directly leveraged by our models and experiments in the following sections.

\section{An architecture for regression problems}
\label{sec:core}

The polynomial regression problem that we tackle is defined by an input pair $(\bfx, \bfd)$ of a domain vector $\bfx = [ \bfx_1; \bfx_2; \cdots; \bfx_N ]\in\mathbb{R}^{DN}$, and a corresponding range vector $\bfd = [ \bfd_1; \bfd_2; \cdots; \bfd_N ]\in\mathbb{R}^{RN}$. The dimensions $D$ and $R$ of $\bfd_i$ and $\bfx_i$ do not have to be the same. For instance, $\bfx_i$ can be an image point ($D=2$) and $\bfd_i$ an intensity value ($R=1$). The relationship between range and domain is assumed to follow a polynomial of given degree, $\bfd_i^\theta = P_{\theta}(\bfx_{i})$, where $\theta$ is the vector of its $M$ coefficients, \eg, $M=6$ for a two-dimensional affine transform. Rewriting this relation as a linear function of $\theta$ reads:
\begin{equation}
\bfd_i^\theta = \bfM_{i}(\bfx_i) \theta,
\label{eq:basic1}
\end{equation}
where $\bfM_i(\bfx_i) \in \mathbb{R}^{R\times M}$ is a design matrix whose structure is maintained across the input data, but whose values are a function of 
the corresponding domain element $\bfx_i$. These design matrices can be stacked into a single matrix $\bfM(\bfx) = [ \bfM_1(\bfx_1); \cdots; \bfM_N(\bfx_N) ] \in \mathbb{R}^{ RN \times M} $ so that: 
\begin{equation}
\bfd^\theta_{\bfx} = \bfM(\bfx) \theta.
\label{eq:basic2}
\end{equation}

Under the assumption that data only undergo additive Gaussian noise, the problem of estimating $\theta$ is reduced to solving:
\begin{equation}
\hat{\theta} = \argmin_\theta \sum_{i=1}^N \norm{\bfd_i - \bfd_i^\theta}_2^2 = \argmin_\theta \norm{\bfd - \bfd_{\bfx}^\theta}_{2}^2,
\label{eq:basic_min}
\end{equation}
from where it follows $\hat{\theta} = \left( \bfM^T \bfM \right)^{-1}\bfM^T \bfd$ (with $\bfx$ hidden for sake of conciseness). This solution corresponds to the simplest possible baseline for polynomial regression, but it is clearly biased under the presence of outliers.

\subsection{Encoder-decoder architecture} To some extent, the problem of outlier removal is similar to the signal denoising problem that stacked denoising autoencoders (SDA)~\cite{vincent2010stacked} address.\footnote{Denoising in SDA is more of a proxy task to facilitate the unsupervised learning of meaningful features from data. However, similar ideas led to a very successful method for image denoising in~\cite{jain2009natural}.} These encoding-decoding architectures, however, are not directly amenable to the polynomial regression problem. 
Indeed, the function that transforms the code into output data should not be learned. It should instead take the form of a fixed, non-trainable differentiable decoding layer. This fixed decoder is simply given by Eq.~\ref{eq:basic2}, \ie, a linear transform of the hypothesized polynomial coefficients $\theta$ based on
a problem-specific design matrix.

\begin{figure}[tb]
	\centering
	\includegraphics[width=.5\linewidth]{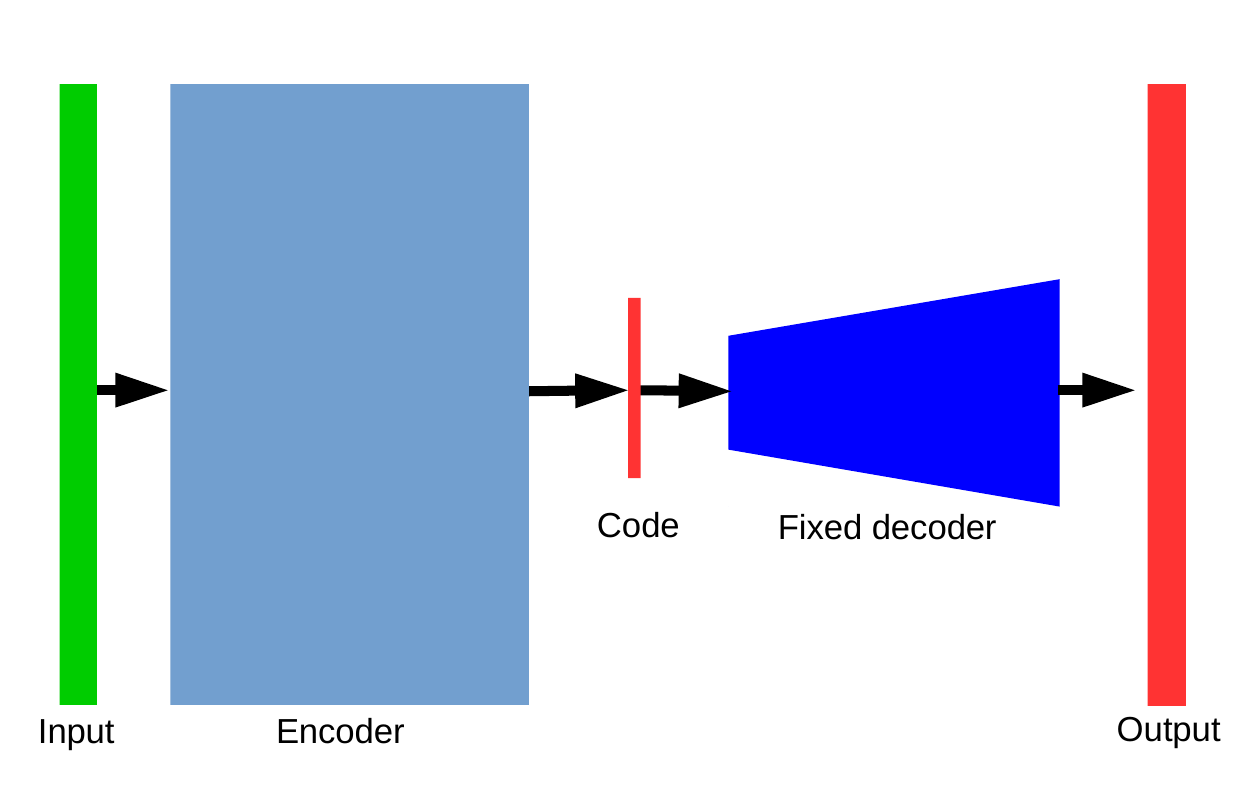}		
	\vspace{-4mm}
	\caption{\scriptsize\label{fig:net}%
		\textbf{Model-based autoencoder with fixed, non-trainable decoder}. The input in green is mapped to a code which effectively becomes the coefficients of a polynomial model when passed through the fixed decoder part (navy blue).}
	\vspace{-4mm}
\end{figure}

Assuming that a learnable encoder, composed of convolution layers, that maps the input data to the polynomial coefficient space is available, an \textbf{Encoder-Fixed Decoder}, or model-based auto-encoder to use the terminology in \cite{tewari2017mofa}, can be formed. Such a network can be trained with well known deep learning training algorithms with a loss function acting on the output (decoded) data. Moreover, the denoising learning trick explained by~\cite{vincent2010stacked} can be readily applied to such architecture, as seen in Fig.~\ref{fig:net}. Granted that training pairs are composed of corrupted and clean data, such a network should be able to learn to regress while ignoring outliers, even structured ones if present in training data.

Furthermore, by means of this training, the ``code'' naturally corresponds to the desired polynomial parameters. An interesting element of this design is that it bypasses the polynomial coefficients themselves at the loss level, eliminating the need for tweaking specific loss functions according to the type of polynomials to be regressed. Indeed, comparing data vectors of the same domain is more straightforward.

\begin{figure}[tb]
	\centering
	\includegraphics[width=.89\linewidth]{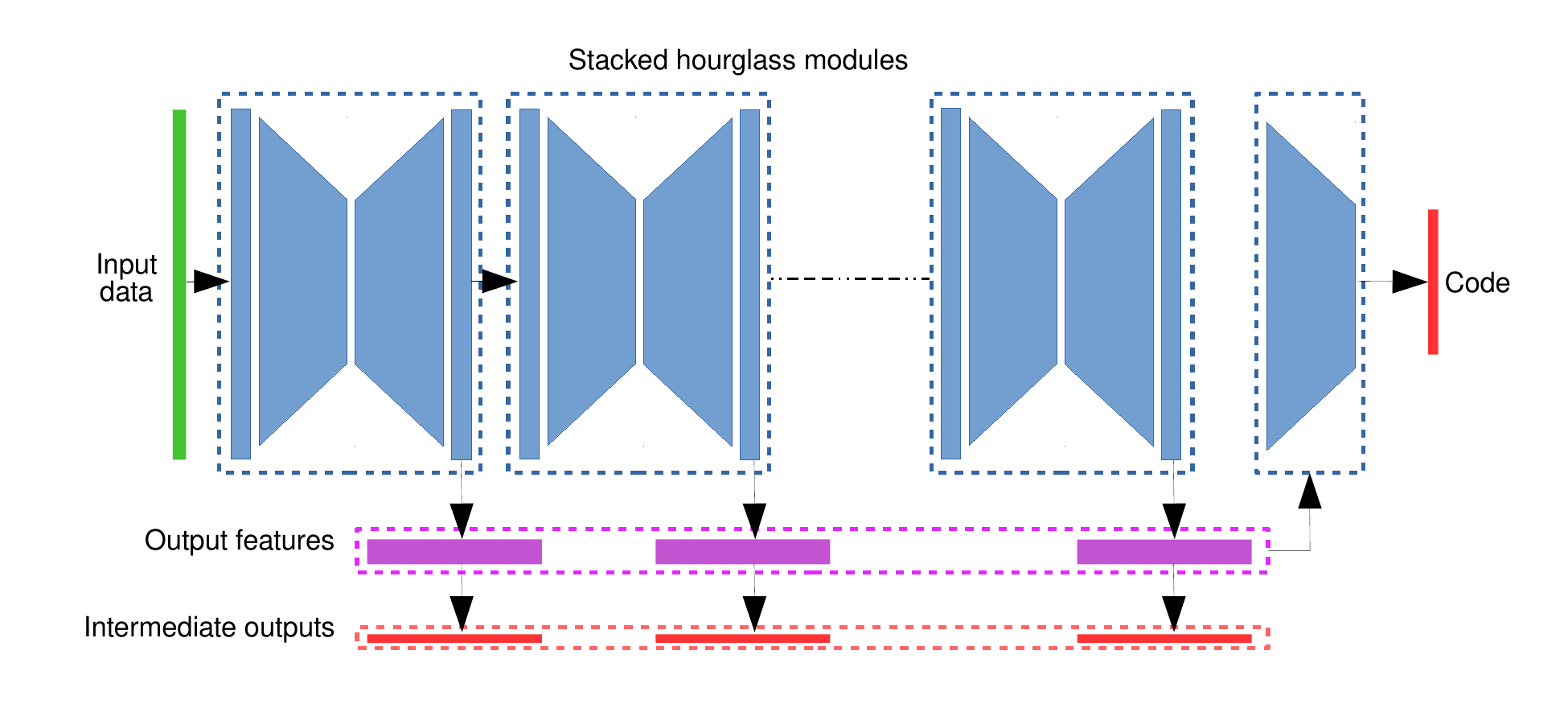}	
	\vspace{-6mm}	
	\caption{\scriptsize\label{fig:encoder}%
		\textbf{Encoder.} The learnable part of our family of networks for deep polynomial regression. The intermediate outputs from each hourglass module are collected for computing the loss together with the final output after the encoding part of our architecture.}
	\vspace{-5mm}
\end{figure}

\subsection{Unsupervised encoder training} Let us, for now, ignore the exact architecture of the encoder part of our family of networks. A common way to train denoising autoencoders was proposed in \cite{vincent2010stacked}, as previously mentioned. This training trick can be categorized as an unsupervised learning method, since pairs of input images and corrupted images are constructed on the fly during training, without the need of human intervention. In the case of polynomial regression, this leaves the door open to fully unsupervised training, as it would be preferred since it is the most common framework to tackle the problem. In our framework, we train our networks by providing randomly generated pairs of clean and corrupted data. The parameters of the random generation process are discussed in supplementary material. Since every sample is generated randomly, training can encompass a very large number of iterations without affecting generalization power of the learned model.

\subsection{Encoder networks} For the encoding part of our family of networks, we propose to use Stacked Hourglass Modules~\cite{newell2016stacked}. Several of the ingredients of SHN seem to be well adapted for the polynomial regression problem with encoder-decoder type of architectures. In particular, the repetitive bottom-up and top-down operations by stacking residual hourglass modules seem to fit the spirit the multi-scale processing spirit of some methods. On the top of that, these residual modules capture scale information, which in the opinion of the authors, it is one of the fundamental elements of problem-tailored regression methods. In the experimental part of this work, we validate these claims by establishing a baseline network composed of more classical feedforward convolutional networks (\ie, purely contractive and without residual connections). As in SHN, we make use of intermediate losses at the output of each hourglass module. The output features are used in a final contractive stage to obtain the polynomial coefficients or ``code'' (See Fig.~\ref{fig:encoder}).

\subsection{Parametric motion model estimation and video stabilization} 
\label{sec:stab}
As previously mentioned, estimation of parametric motion models is a very good example of a polynomial regression problem with naturally strong outliers. In such a setting a polynomial motion model for a moving scene is interpreted as the dominant scene motion stemming from camera motion. In that sense, outliers correspond to moving foreground objects, which can occupy large areas of the scene.

A common way to perform video stabilization is to compute a temporally and spatially smooth optical flow map~\cite{liu2014steadyflow}. One way to achieve this is to compute at each instant a bi-dimensional optical flow ($R=2$) over the image domain ($D=2$) and then fit a polynomial function to it.
Given an input optical flow map $\mathcal{V} = \{ \mathbf{f_\bfx} \}_{\bfx\in\Omega}$, one can fit a polynomial function $\mathbf{f}^\theta $ computable at every position $\bfx=(x_1,x_2)$ of the image grid $\Omega$, so that:
\vspace{-1mm}
\begin{equation}
\mathbf{f}_\bfx^\theta =
\begin{bmatrix} u_\bfx^\theta\\v_\bfx^\theta \end{bmatrix} = \mathbf{M}(\bfx) \theta,
\label{eq:basic2d}
\end{equation}

\noindent where, $\theta$ is a column vector containing the parameters of a polynomial motion model. Let us consider, for sake of generality, full quadratic motion models with twelve coefficients ($M=12$).\footnote{Any other common motion models like 4-parameter affine, and 8-parameter (corresponding to rigid motion of a planar scene) could be considered as well.} Then, the matrix $\textbf{M}(\bfx)$ in Eq.~\ref{eq:basic2d} takes the form:
\vspace{-1mm}
\begin{equation}
\vspace{-1mm}
\setcounter{MaxMatrixCols}{20}
\mathbf{M}(\bfx) =
\begin{bmatrix} 1 & 0 & x_1 & x_2 & 0   & 0   & x_1^2 & x_1 x_2 & x_2^2 & 0     & 0       & 0 \\
0 & 1 & 0   & 0   & x_1 & x_2 & 0     & 0       & 0     & x_1^2 & x_1 x_2 & x_2^2
\end{bmatrix},
\label{eq:motion2d_M}
\end{equation}

\noindent and $\theta \in \mathbb{R}^{12}$. Equations~\ref{eq:basic2d} and~\ref{eq:motion2d_M} are specializations of previously introduced Eq.~\ref{eq:basic1} for a polynomial function with two-dimensional domain. This means that the proposed network of Fig.~\ref{fig:net} applies directly to this problem. We explain data generation and training issues in the supplementary material. In Section~\ref{sec:experiments}, we report experimental results and a direct application for the problem of video stabilization.

\section{Experimentas}
\label{sec:experiments}

\begin{table}[tb]
	\centering 
	\scriptsize	
	\caption{\scriptsize\textbf{Regressing scalar functions with 4-th degree polynomial of one variable} Results for 6 different testing datasets at increasingly higher outlier ratios with a fixed noise standard deviation of $0.01$. The numbers are the mean squared error between generated clean data and the outputs of respective methods. The lower the number, the better the accuracy. We indicate in bold best results and underline second best for every column.}
	\begin{tabular}{c|@{\hskip 0.1 in}ccccccc}
		
		\hline
		& \multicolumn{7}{c}{Outlier ratio} \\
		& $0\%$ & $10\%$ & $20\%$ & $30\%$ & $40\%$ & $50\%$ &  Average\\
		\hhline{========}
		LSE           & \textbf{1.9e-6} & 6.142 & 16.92 & 37.63 & 53.38 & 74.60 & 31.44\\
		RANSAC        & 0.106 & 0.133 & 3.775 & 2.414 & 12.04 & 27.87 & 7.720\\
		IRWLS         & 0.328 & 1.120 & 2.917 & 5.574 & 22.10 & 32.26 & 10.71\\
		\hhline{========}
		HalfNet Data 1 & 12.95 & 32.11 & 45.39 & 64.69 & 77.50 & 87.21 & 53.30\\
		HalfNet Data 1\&2 & 0.222 & 0.341 & 0.912 & 1.147 & 2.442 & 4.942 & 1.667\\
		HalfNet Data 1+2 & 0.191 & 0.339 & 0.901 & 1.148 & 2.157 & 5.265 & 1.668\\
		\hhline{========}		
		FullNet w.o. D. & 0.219 & 0.529 & 0.634 & 0.839 & 1.250 & 2.651 & 1.020\\
		FullNet R.L. & 0.110 & 5.966 & 12.543 & 20.815 & 36.464 & 52.935 & 21.47\\
		FullNet Data 1 & 0.107 & 19.65 & 33.12 & 39.65 & 48.65 & 59.33 & 3.341\\
		FullNet Data 1\&2 & \underline{0.033} & \underline{0.111} & \underline{0.178} & \textbf{0.239} & \textbf{0.364} & \textbf{0.968} & \textbf{0.315}\\
		FullNet Data 1+2 & 0.074 & \textbf{0.086} & \textbf{0.163} & \underline{0.258} & \underline{0.458} & \underline{0.989} & \underline{0.338}\\
		\hline
		
	\end{tabular}
	\vspace{-4mm}
	\label{tab:expe1}
\end{table}

We present first two sets of experiments in Section~\ref{sec:experiments1}. Their goal 
is to validate the design decisions explained in previous sections, and demonstrate that they hold even when the dimensionality of the problem changes. We start with least squares (LSE) as the simpler baseline. We also provide results with three conventional robust algorithms, RANSAC, and a robust Tukey~\cite{huber2011robust} estimator solved with the Iterative Re-weighted Least Squares (IRWLS).
Finally, in Section~\ref{sec:experiments2}, we present a simple video stabilization pipeline based on parametric motion models regressed from optical flow maps.

\subsection{Deep polynomial regression}
\label{sec:experiments1}

\paragraph{Regression of scalar polynomials.} We start with a toy experiment consisting of a simple $1D$ regression problem ($R=1$ and $D=1$) with scalar polynomials of degree four. 
 The evaluation of our framework for deep polynomial regression is split into two types of network. First, the ``Half-Nets" refer to our Encoder-Decoder networks without stacked hourglass modules. Instead, ``Half-Nets" are composed only of contractive convolutions for the encoder part, and our fixed decoder on top of it. Furthermore, we want to determine if split training~\cite{ilg2017flownet} presents any advantage over training with a single complex dataset. Thus, \textit{Half-Net Data 1} is trained only on a dataset encompassing functions with smaller magnitude, contaminated with little noise and with only few outliers. \textit{Half-Net Data 1\&2} is refined on a second stage with a more complicated dataset encompassing a larger variety of generation modes, more noise and up to $33\%$ structured outliers. During evaluation outlier ratios up to $50\%$ are tested. Later on, \textit{Half-Net Data 1+2} is a network trained with a dataset encompassing both the simple and complicated generation schemes combined in a single stage.

A second type of networks encompassing SHN for the encoder part are referred as ``Full-Nets''. In the same order as for the ``Half-Nets'', \textit{Full-Net Data 1}, \textit{Full-Net Data 1\&2}, \textit{Full-Net Data 1+2} study the importance of split training. In order to determine the value of our fixed decoder as part of the network, we also train \textit{Full-Net w.o. D} with a mean square loss on the produced parametric coefficients. Finally, \textit{Full-Net R.L.} is trained with a robust loss. For \textit{Full-Net R.L.}, instead of training with the proposed dataset composed of pairs of clean and contaminated vectors, we train only with contaminated data\footnote{The idea of this particular experiment is assessing if only by means of a robust loss, neural networks are able to learn to reject outliers.}. The robust loss in this network is expected to reject outliers automatically as it is proposed in~\cite{belagiannis2015robust}. Both \textit{Full-Net w.o. D} and \textit{Full-Net R.L.} are trained on the complex dataset (\textit{Data 1+2}). All the results can be observed in Table~\ref{tab:expe1}.

A first conclusion that can be drawn from Table~\ref{tab:expe1} is that, overall, neural-based regression is more stable than classical robust methods (RANSAC or IRLWS) in the face of outlier corruption. Another interesting outcome of our experimentation is that for scalar polynomial regression, split training does not seem to provide a large gain in terms of prediction error. However, it does achieve the absolute best results in comparison to other dataset configurations. Perhaps more interestingly, all of our networks with a preceding hourglass module present overall better results than any other set-up. In particular, \textit{Full-Net Data 1\&2} seems to achieve lowest errors, except for the corruption-free case, where LSE is the optimal estimator, and therefore expected. An interesting finding is that our denoising training effectively teaches our networks how to be robust to outliers. In contrast, when simply training with a robust function (\textit{Full-Net R.L.}) results tend to be poor. This enforces the notion that \textbf{neural networks learn how to be robust more easily by example than by application of robust losses}. Finally, it is worth mentioning that training through our hard-wired decoder does indeed offer a jump in accuracy of the regression. This interesting fact seems to be in agreement to very recent findings in neural estimation of image geometry~\cite{detone2017superpoint}.

\paragraph{Regression of 2D polynomials (parametric motion models).}

\begin{table}[tb]
	\centering 
	\scriptsize
	\caption{\scriptsize\textbf{Regressing vector fields with 2-nd degree polynomial of two variables.} Results for 6 different testing datasets at increasingly higher outlier ratios with a fixed noise standard deviation of $0.5$. The error values with the Euclidean norm between generated clean data and the outputs of respective methods. We bold best results and underline second best for every column.}
	\begin{tabular}{c|@{\hskip 0.1 in}ccccccc}
		
		\hline
		& \multicolumn{7}{c}{Outlier ratio} \\		
		& $0\%$ & $10\%$ & $20\%$ & $30\%$ & $40\%$ & $50\%$ & Average \\
		\hhline{========}
		LSE           & \textbf{2.0e-5} & 0.047 & 0.054 & 0.078 & 0.093 & 0.114 & 0.0643\\
		RANSAC        & 0.015 & 0.019 & 0.022 & 0.046 & 0.068 & 0.093 & 0.0438\\ 
		IRWLS         & 0.002 & 0.023 & 0.061 & 0.067 & 0.077 & 0.110 & 0.0566\\ 
		\hhline{========}
		HalfNet Data 1 & 0.021 & 0.037 & 0.038 & 0.041 & 0.048 & 0.060 & 0.0408\\ 
		HalfNet Data 1\&2 & 0.019 & 0.032 & 0.036 & 0.040 & 0.051 & 0.038 & 0.0360\\
		HalfNet Data 1+2 & 0.018 & 0.030 & 0.033 & 0.041 & 0.052 & 0.054 & 0.0380\\
		\hhline{========}
		FullNet w.o.D. & 0.009 & 0.008 & 0.012 & 0.022 & 0.058 & 0.077 & 0.0310    \\
		FullNet R.L. & 0.019 & 0.027 & 0.046 & 0.059 & 0.067 & 0.235 & 0.0755 \\
		FullNet Data 1 & 0.011 & 0.024 & 0.035 & 0.039 & 0.042 & 0.046 & 0.0328\\
		FullNet Data 1\&2 & \underline{0.006} & \underline{0.007} & \underline{0.009} & \underline{0.012} & \underline{0.015} & \underline{0.022} & \underline{0.0118}\\
		FullNet Data 1+2 & \underline{0.006} & \textbf{0.006} & \textbf{0.007} & \textbf{0.010} & \textbf{0.013} & \textbf{0.019} & \textbf{0.0101}\\
		\hline
		
	\end{tabular}
	\vspace{-2mm}
	\label{tab:expe2}
\end{table}

To show that our findings hold for several types of polynomial regression problems, we repeat the previous experiments, this time for a polynomial regression problem of higher dimensionality ($R=2$, $D=2$), involving the full quadratic motion model of~Eq.~\ref{eq:motion2d_M} ($M=12$). The experiments are collected in Table~\ref{tab:expe2}, where the items have the same meaning than for Table~\ref{tab:expe1}. From Table~\ref{tab:expe2}, we observe that most of the findings for scalar regression regression hold for the vector field case. This time, \textit{Full-Net Data 1+2} delivers the best overall results, with a MSE score that is roughly three times lower than the baseline without fixed decoder (\textit{FullNet w.o.D.}) and almost four times lower than the best configuration of networks without stacked hourglass modules (\textit{HalfNet Data 1\&2}). In this set-up \textit{Full-Net Data 1+2} and \textit{Full-Net Data 1\&2} do not seem to present significantly different results. Finally, it is worth mentioning that our training procedure presents training samples with at most $30\%$ outlier ratio, while the experiments reach up to $50\%$. Interestingly, our best models generalized very well for the extreme outlier conditions, achieving errors of one order of magnitude better than classic baselines for the vector field case, and two orders of magnitude for the scalar function case.

\subsection{Dominant motion estimation and video stabilization}
\label{sec:experiments2}

We now report experiments to demonstrate that the proposed approach is relevant for real data, even if training is conducted on synthetic data only and without supervision. In particular, we take the best network from the second set of experiments in Section~\ref{sec:experiments1} and use it directly to compute parametric motion models from optical flow. We start from the popular optical flow baseline method by Brox \etal~\cite{brox2004high}.\footnote{It should be noted that one could replace the off-the-shelf optical flow method by a more precise one. Furthermore, one could connect a deep-learning based method to ours, enabling end-to-end training from images. We leave this for future research.} Robustly fitting a quadratic motion model to an estimated complex non parametric flow is very challenging due to flow inaccuracies (which can be slightly structured) and, more importantly, to highly structured outliers that foreground objects induce.

\begin{figure*}[t]
	\centering
	\begin{overpic}[width=0.80\textwidth]{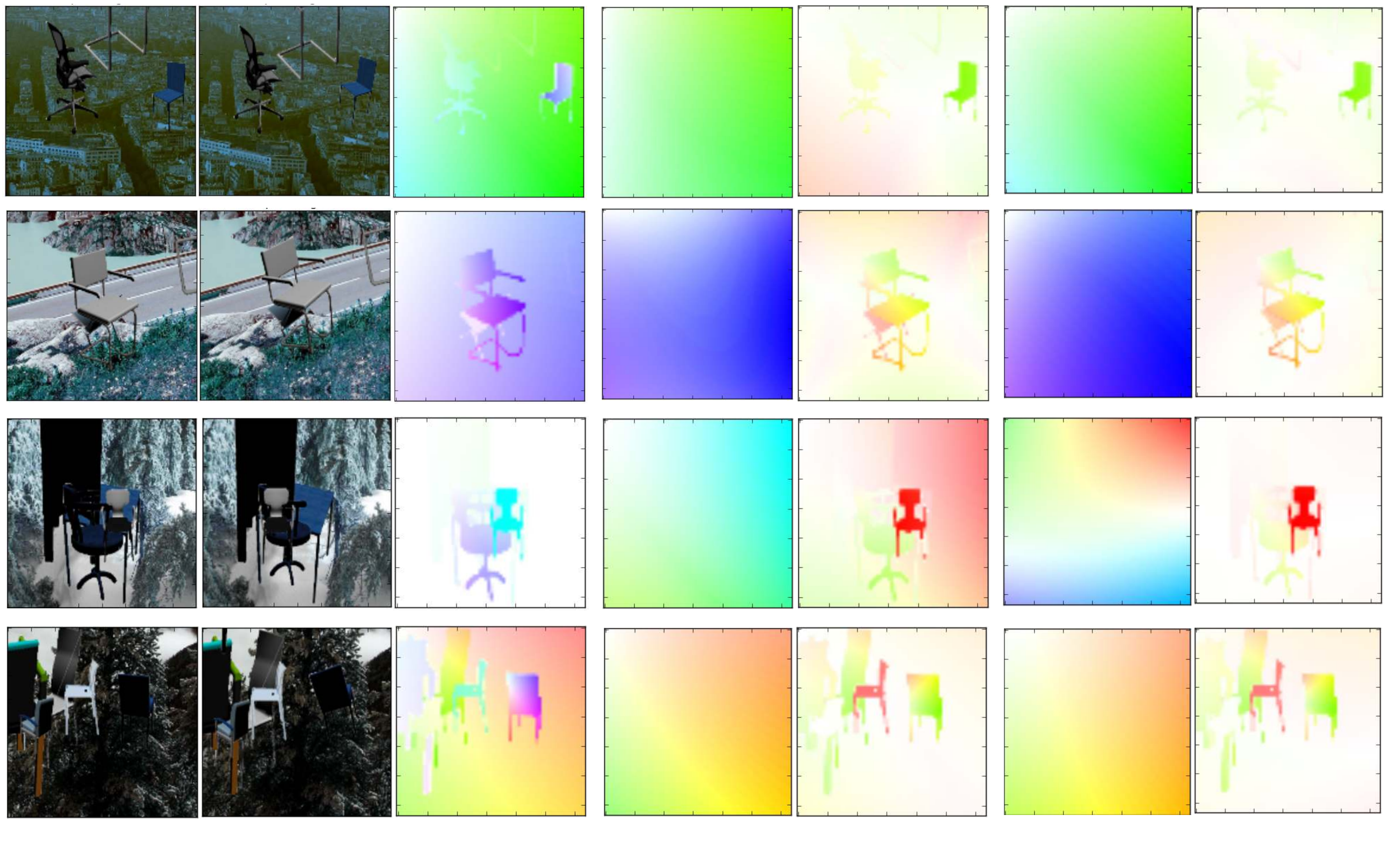}	
		\put (7,1) {(a)}
		\put (20,1) {(b)}
		\put (34,1) {(c)}
		\put (49.5,1) {(d)}
		\put (63,1) {(e)}
		\put (77.5,1) {(f)}
		\put (92,1) {(g)}
	\end{overpic}
	\vspace{-2mm}
	\caption{\scriptsize\textbf{Visual results on the FlyingChairs dataset.} A different scene is shown in each row. (a-b) The image pair; (c) the input optical flow map (ground-truth); (d) output of \textit{FullNet Data 1}; (f) The corresponding results for \textit{FullNet Data 1+2}. Observe the simplification of the input optical flow in a parametric motion model that gets rid of outliers (motion of the chairs); (e) and (g): normalized per-pixel difference between the output and input flows corresponding to our model trained on the first and second datasets, respectively. Observe that the larger differences (more saturated colors) are, generally, in the pixels depicting a moving chair.
	}
	\vspace{-2mm}
	\label{fig:flying}
\end{figure*}

\begin{figure*}[t]
	\centering
	\includegraphics[width=0.98\textwidth]{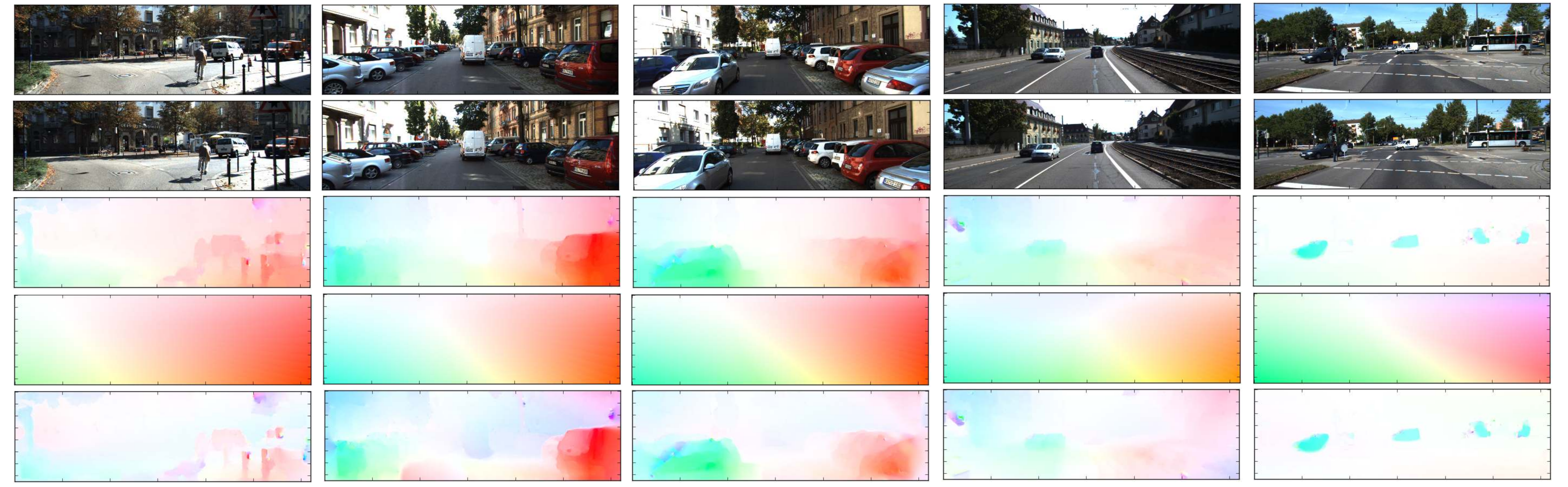}	
	
	\vspace{-2mm}
	\caption{\scriptsize\textbf{Visual results on the Kitti dataset.} A different scene is shown in each column: (top) input image pair; (third row) input optical flow; (fourth row) quadratic motion field obtained by our deep regression network; (fifth row) pixel-wise difference between inputs and outputs. 
	\vspace{-2mm}
	}
	\label{fig:kitti}
\end{figure*}

We first start by analyzing the behavior of our model on a synthetic optical flow dataset, namely~\textit{FlyingChairs} in Fig.~\ref{fig:flying}. This dataset is composed of synthetically rendered chairs composed onto moving backgrounds. Our model captures what can be interpreted as dominant scene motion, mostly corresponding to the background. Interestingly, the high magnitude values of the difference between the input and output flow maps correspond to foreground objects, \ie, the moving chairs. In this sense, our method robustly captures global motion, showing high insensitivity to outliers. In Fig.~\ref{fig:flying}, we can also see the difference between the same model trained on two different datasets (columns e and g). Evidently,~\textit{FullNet Data 1+2}, being trained on a more complex outlier-generation setting, better captures global motion than \textit{FullNet Data 1}.

In a more realistic set-up, we take the Kitti dataset and compute optical flow maps between pairs of frames with~\cite{brox2004high}. From these maps, we fit quadratic motion models with our best network, resulting in the visual results in Fig.~\ref{fig:kitti}. The reader should notice that the simplified flow maps stemming from the fitted quadratic motion models conform well with the dominant ego-motion induced on the scene by the displacement of the embarked camera.

  \begin{figure}[tbh]
  	\centering
  	\begin{center}
  		\setlength{\tabcolsep}{2pt}
  		\begin{tabular}{cccc}
  			\includegraphics[width=0.22\linewidth]{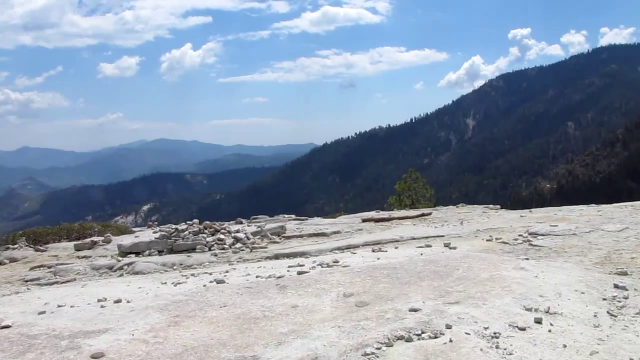} &
  			\includegraphics[width=0.22\linewidth]{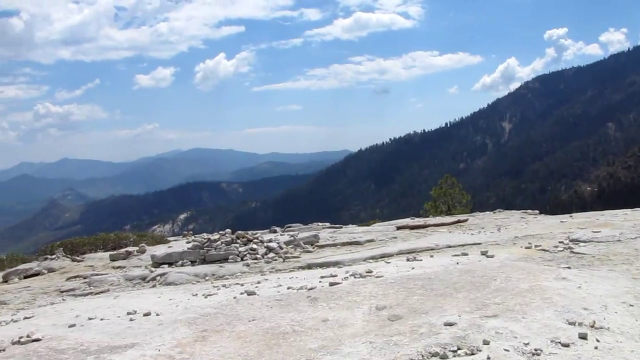} &
  			\includegraphics[width=0.22\linewidth]{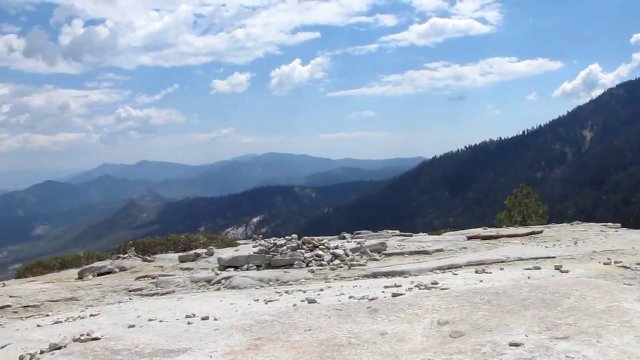} &	  			
  			\includegraphics[width=0.22\linewidth]{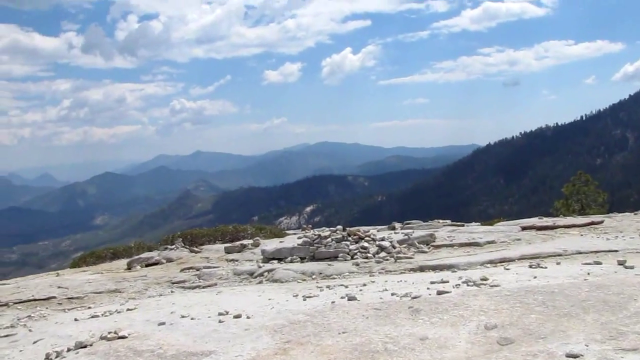} \\

  			\includegraphics[width=0.22\linewidth]{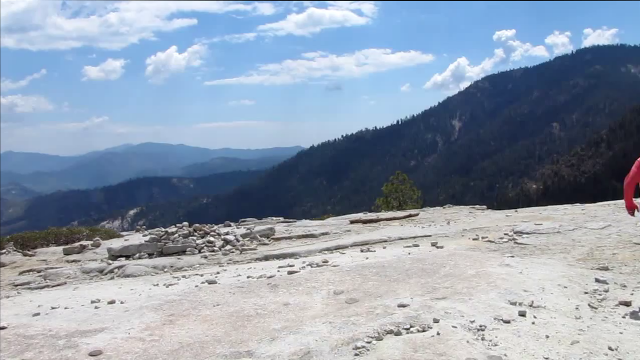} &
  			\includegraphics[width=0.22\linewidth]{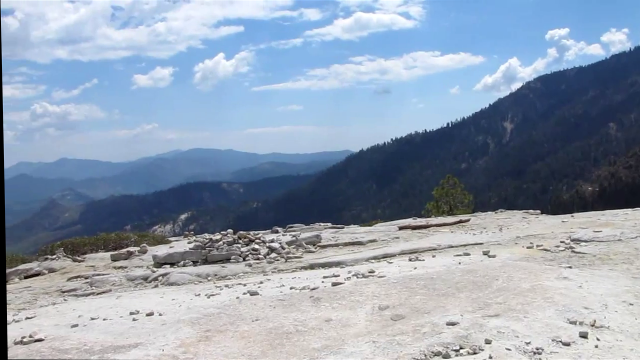} &
  			\includegraphics[width=0.22\linewidth]{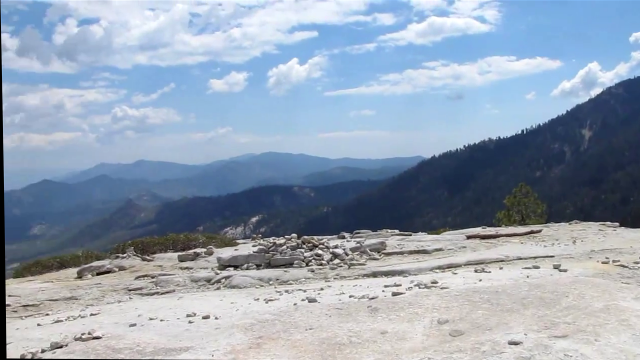} &	  			
  			\includegraphics[width=0.22\linewidth]{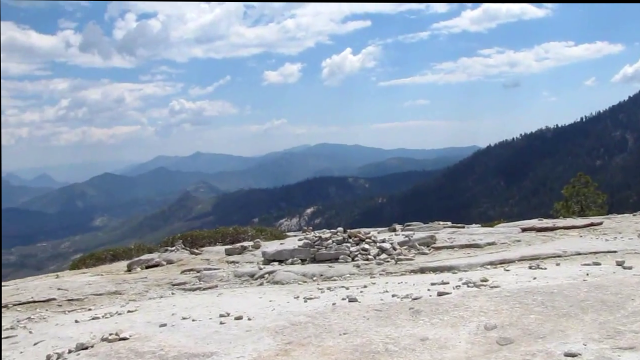} \\

  		\end{tabular}
  	\end{center}
   	\vspace{-2mm}
  	\caption{\scriptsize\textbf{Video stabilization results}. Estimated quadratic dominant motion in a real scene is used to backwarp images aiming at video stabilization. First row: original unstable input. Second row: stabilized images. The out-of-frame holes (black) are left so the reader appreciates better the motion compensation.}
  	\label{fig:stab}
		\vspace{-2mm}
  \end{figure}
  
  \paragraph{Video stabilization.} A common application of dominant motion estimation is video stabilization: cancelling the higher frequencies of background visual motion indeed requires an accurate estimation of this flow field. In Fig.~\ref{fig:stab} we show stabilization results with out deep polynomial regression method on a sequence undergoing camera-shaking by effect of strong wind. Observe that the image sequences are correctly warped to compensate for dominant motion.
  In a further improvement, we smooth pixel profiles in a similar fashion as~\cite{liu2014steadyflow} to achieve better temporal smoothness. Our algorithm delivers high quality results, considering that we did not introduce any higher level considerations for video stabilization problem.

\section{Model details}
\label{sec:model}

In the following paragraphs we give important details of the models used during the experimental part of our work.

\subsection{Scalar function regression}
\label{sec:scalarmodel}
The problem we tackle in Section ~\ref{sec:experiments2} is the regression of 4th degree polynomial coefficients. This set-up is amenable to the multi-scale, repetitive processing of stacked hourglass networks. However, we replace all the 2D convolutions by 1D convolutions in account of the input data structure.

\paragraph{Hourglass modules.} Each convolutional layer in the reductive part of an hourglass module is formed by 32 convolutional kernels of size three. At the output of each one of these convolutional layers a sequence of three operations is applied. These are, ReLU, 1D max pooling with stride 2, and 1D batch normalization. Furthermore, after the ReLU operations a forward skip connections through convolutions of the same size are connected to bilinear upsampling operations in the enlarging part of the hourglass modules. For this experiment we found that stacking more than two hourglass modules did not improve performance much further.

\paragraph{Model-based autoencoder.} 3 layers of 1D convolutional operation of kernel size 3 are connected to the concatenated output features of the stacked hourglass networks. Each convolution is followed by 1D batch normalization and ReLU operations. A final convolution layer of kernel size 1 is connected to reduce the depth from 96 planes to only 8. From these, a fully connected layer without non-linearity delivers the 
polynomial parameters, that are later brought back to the input data space through our fixed polynomial decoder.

\subsection{Vector field regression}
\label{sec:vectormodel}

We now explain the model used for the experiments of Section~\ref{sec:experiments2}. Aligned with the intention of the paper of providing a general model for regression problems, we use a model that is identical to the one used for scalar regression with only a few minor differences. Thus, each 1D convolution of previous model is replaced by a 2D convolution of twice as many planes. Furthermore, the fixed decoder of the model-based autoencoder is replaced to match the new polynomial operation.

\section{Dataset generation}
\label{sec:dataset}

\begin{figure*}[t]
	\centering
	\includegraphics[width=0.98\textwidth]{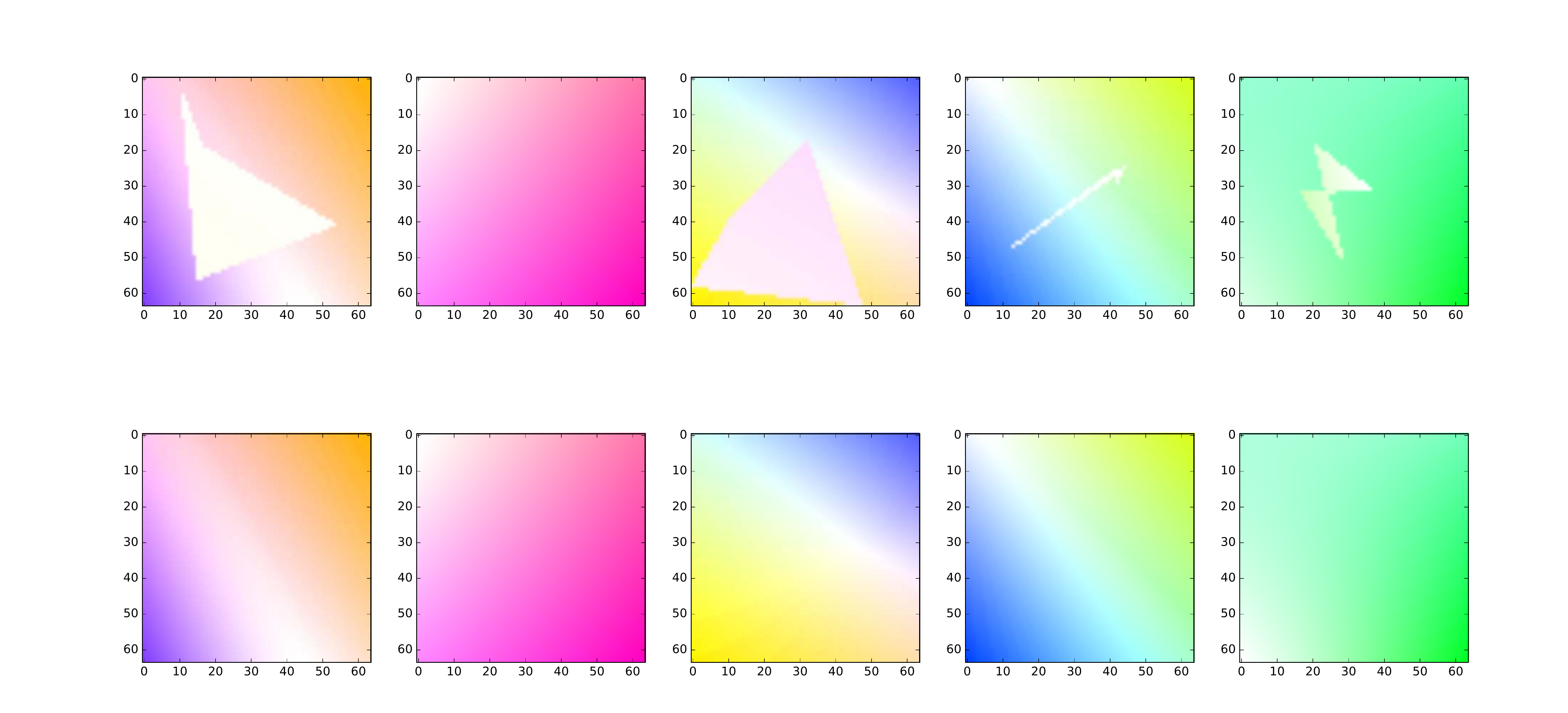}	
	
	\vspace{-2mm}
	\caption{\scriptsize\textbf{Randomly generated training pairs.} Top row: Contaminated input. Bottom row: clean output.
	\vspace{-2mm}
	}
	\label{fig:data}
\end{figure*}

Two dataset generators are used for training of our regression models, namely \textit{Data 1} and \textit{Data 2}. Each one of them is composed by pairs of ``input'' and ``output'' samples. The ``output'' samples are clean arrays of values stemming from a polynomial model, while the ``input'' samples are corresponding contaminated maps. The pair simulates a supervised training pair, but it is generated online randomly. See Fig.~\ref{fig:data} for an illustration of data generated for the vector field case.

The main difference between~\textit{Data 1}~and \textit{Data 2} is the intensity of the contamination. The idea is that~\textit{Data 1} would allow neural networks to learn the basic operations for regression without many distractions, while a subsequent dataset would refine the network for handling more complex contamination. Thus, we set the maximum outlier ratio for~\textit{Data 1} to 0.1, while we set it to 0.3 for~\textit{Data 2}. Gaussian noise is set to 0.1 for~\textit{Data 1} and 0.5 for~\textit{Data 2}. The structured outliers correspond to randomly selected polygonal supports encompassing another randomly sampled polynomial model.

\section{Conclusion}
\label{sec:conclusions}

In this paper we have proposed a neural approach to robust polynomial regression. We carried out an experimental study spanning several important hints from previous state-of-the-art. We have provided a general class of architectures that learn to deal with outliers effectively. In particular, the spatially-consistent outliers that have been an important problem for classical polynomial regression methods can be handled by properly trained and designed neural nets. Moreover, we have shown that these findings hold for different settings of polynomial regression. The proposed architecture captures two important common design strategies for polynomial regression: spatial awareness (effective through convolutions), and multi-scale processing (by stacked hourglass modules). This design, in conjunction with a stacked denoising autoencoder training with simulated outliers, results in models that robustly learn how to handle largely contaminated data. Furthermore, networks trained with purely synthetic data are able to generalize to real data, leading to very good accuracy for global motion estimation and effective use for video stabilization. We can expect that our findings can be generalized to other types of regression problems.

\bibliographystyle{splncs}
\bibliography{tex/biblio}

\end{document}